\documentclass[letterpaper, 10 pt, conference]{ieeeconf}  % Comment this line out if you need a4paper

\IEEEoverridecommandlockouts                              % This command is only needed if 
\usepackage{graphics} % for pdf, bitmapped graphics files
\usepackage{epsfig} % for postscript graphics files
\usepackage{mathptmx} % assumes new font selection scheme installed
\usepackage{times} % assumes new font selection scheme installed
\usepackage{amsmath} % assumes amsmath package installed
\usepackage{amssymb}  % assumes amsmath package installed
\makeatletter
\@ifundefined{labelindent}{}{%
  \let\labelindent\relax
  
}
\makeatother
\usepackage{enumitem}
\usepackage{tablefootnote}
\usepackage{subcaption}
\usepackage{multirow}
\usepackage{siunitx}
\usepackage{booktabs}
\usepackage{soul}
\usepackage[table]{xcolor}  
\usepackage{cite}
\usepackage{hyperref}
\usepackage[capitalize]{cleveref}

\newcommand{\etal}{\textit{et al.}}
\newcommand{\ie}{\textit{i}.\textit{e}., }

\title{\LARGE \bf
TactSpace: Learning a Physics-enriched Shared Latent Space \\ for Tactile Sim-to-Real Transfer
}

\author{Arunim Joarder$^1$, Arjun Bhardwaj$^1$, Ren\'{e} Zurbr\"{u}gg$^{1,3}$, Mayank Mittal$^{1,4}$, \\
Florin Püntener$^2$, Sira Bielefeldt$^2$, Cosmin Roman$^2$, Vaishakh Patil$^1$, Marco Hutter$^1$ % <-this % stops a space
\thanks{This work was supported by an ETH Zurich Research Grant No. 22-2 ETH-47, Swiss National Science Foundation (NCCR Automation grant no. 51NF40 225155) and the ETH AI Center. $^1$Robotic Systems Lab, ETH Z\"{u}rich; $^2$Micro- and Nanosystems Lab, ETH Z\"{u}rich; $^3$ ETH AI Center; $^4$NVIDIA.}
\thanks{Email: {\tt\small \{ajoarder,abhardwaj,zrene\}@ethz.ch}}%
}
\renewcommand{\baselinestretch}{0.9865}

\begin{document}
\bstctlcite{IEEEexample:BSTcontrol}

\maketitle
\thispagestyle{empty}
\pagestyle{empty}

\begingroup
\let\clearpage\relax
\begin{abstract}

Tactile sensing provides direct measurements of contact interactions that are essential for robotic manipulation. However, current simulators lack the fidelity to faithfully model the complex deformation and transduction mechanics of tactile sensors, severely hindering sim-to-real transfer in robot learning pipelines. To address this challenge, we propose a multi-modal representation learning framework that aligns heterogeneous tactile modalities within a shared latent space, eliminating the need for accurate raw-signal simulation while preserving relevant contact information. Our approach employs modality-specific encoders to project diverse tactile observations, such as simulated penetration depth and real-world capacitance, into a common embedding space. The model is trained using self- and cross-reconstruction objectives alongside contrastive alignment, encouraging modality-invariant yet information-rich representations. We evaluate the learned embeddings on indenter shape identification, force prediction, and geometric reconstruction tasks, training exclusively in simulation and testing directly on real sensor measurements. Our results demonstrate zero-shot sim-to-real transfer across physically dissimilar representations. Furthermore, incorporating multi-physics simulation modalities yields more informative embeddings that transfer across diverse downstream tasks, demonstrating a 16.7\% reduction in force prediction error and a 45.8\% reduction in shape reconstruction error. Finally, we release an efficient Warp-based implementation of a penalty-based tactile simulation model for Isaac Lab, enabling scalable tactile data generation. For videos of the data collection and additional supplementary materials, please refer to the project
website: \href{https://leggedrobotics.github.io/tactspace-web/}{https://leggedrobotics.github.io/tactspace-web/}.

\end{abstract}
\section{INTRODUCTION}

Tactile sensing is fundamental to contact-rich robotic manipulation, providing direct measurements of surface deformation and contact forces. Consequently, a wide range of tactile sensors, ranging from high-resolution vision-based skins~\cite{Lambeta_2020, s17122762} to robust capacitive~\cite{liu2022printed, johannes_tactful} and piezoresistive arrays~\cite{9636677}, has been deployed for tasks spanning slip detection~\cite{veiga2018grip}, texture recognition~\cite{CAO2024104688, 9341333}, and material classification. Despite its importance, learning-based tactile manipulation remains bottlenecked by the scarcity of large-scale, contact-rich interaction data~\cite{luo2025tactileroboticsoutlook}. 

Collecting tactile data on hardware is slow, costly, and unsafe, making simulation a necessary component of scalable tactile learning. While sim-to-real methods have become a standard paradigm for training robotic policies~\cite{yang2024anyrotate, qi2023general}, incorporating tactile sensing into this pipeline remains challenging. Prior works attempt to bridge the sim-to-real gap by developing high-fidelity, sensor-specific simulations~\cite{si2021taximexamplebasedsimulationmodel, zhao2024fots}. However, such pipelines rely heavily on modeling assumptions, require substantial engineering effort, and do not readily extend to other types of tactile sensors, such as capacitive sensors, whose signals arise from complex electromechanical and material interactions.

Furthermore, modern robot learning workflows often rely on hardware-accelerated physics-based simulators, that allow rapid, large-scale data generation. However, this scalability comes at the cost of physical accuracy: rigid-body simulators currently lack the ability to model complex material deformation, detailed contact mechanics, and realistic force propagation. As a result, the geometric proxies provided by these simulators differ substantially from real tactile sensor outputs. Consequently, when control policies are trained directly on these simulated tactile proxies, the severe domain gap prevents successful transfer to real-world hardware.

\begin{figure}
    \centering
    \includegraphics[width=0.95\linewidth]{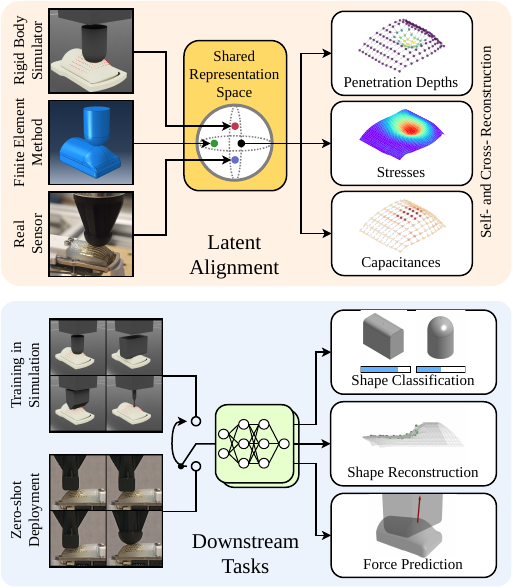}
    \caption{(Top) Multi-modal tactile inputs are aligned into a shared latent space, enabling self- and cross-modal reconstruction. (Bottom) A model trained exclusively on simulated embeddings generalizes zero-shot to real sensor data across downstream tasks including shape classification, reconstruction, and force prediction.}
    \label{fig:teaser}
    \vspace{-16pt}
\end{figure}

% Representation learning provides a powerful mechanism for bridging diverse sensory domains, mapping heterogeneous signals into a shared embedding space that captures their underlying structure while abstracting away modality-specific details~\cite{8715409}. For example, models such as CLIP~\cite{radford2021learningtransferablevisualmodels} learn joint representations for images and text by enforcing cross-modal consistency rather than reproducing raw inputs. We hypothesize that contact interactions admit a structured, modality-agnostic latent representation encoding intrinsic geometric and physical regularities. If simulated and real tactile observations are aligned within such a latent space, sim-to-real transfer need not depend on matching sensor-specific raw signals in the underlying simulation.
%\mayank{In this work, we look specifically at capacitance-based tactile sensor~\cite{tactful sensor}. The sensor has a gel membrane which is complex and non-linearly deformable. Modeling these effects are not possible in the current simulators. Showing it for a complex prototype like this, is nie because other modalities should also follow. -- Phrasing it roughly in this spriti to make it clear that we look at this sensor type and we do it for this because it is hard than other gelsight things that exist.}

% Also, just using real-data does not give us any dense contact information, which is provided by the simulation. But the simulation latents are not transferable.
In this paper, we propose an autoencoder-based framework that explicitly constructs a shared representation between simulated observations and sensor measurements. Trained via self- and cross-reconstruction objectives alongside an InfoNCE contrastive~\cite{oord2019representationlearningcontrastivepredictive} alignment, our architecture ensures that corresponding physical interactions across domains produce similar embeddings, successfully abstracting away sensor-specific artifacts. A central advantage of our approach is the ability to integrate complementary simulation modalities that capture different levels of physical abstraction. We jointly leverage scalable contact geometry and kinematics from rigid-body simulators (NVIDIA Isaac Lab~\cite{mittal2025isaaclab}) alongside the detailed stress fields and dynamic material responses modeled by finite-element analysis (FEA) solvers (ABAQUS~\cite{0b112d0e5eba4b7f9768cfe1d818872e}). This constructs a latent space that encodes a richer contact phenomena than either modality may achieve on their own. We evaluate the learned representation by assessing sim-to-real transfer performance on downstream contact-rich prediction tasks, shown in~\cref{fig:teaser}. In addition, we release a tactile simulation plugin for Isaac Lab for scalable collection of contact geometry and force estimates within standard robot manipulation pipelines.

In summary, our main contributions are:
\begin{itemize}
    \item A representation-learning framework for tactile sim-to-real transfer that aligns heterogeneous modalities within a shared latent space and allows for the zero-shot transfer of representations to raw sensor measurements,
    \item A multimodal encoder--decoder framework that integrates cross-reconstruction with contrastive alignment to learn modality-invariant representations, improving performance with 16.7\% reduction in force prediction error and 45.8\% reduction in shape reconstruction error,
    \item A tactile simulation plugin for Isaac Lab that facilitates scalable tactile data generation and seamless integration into modern robot learning workflows.
\end{itemize}

\section{RELATED WORK}

\subsection{Tactile Simulation and Sim-to-Real Transfer}

Scalable tactile simulation has become increasingly essential for training robot learning algorithms~\cite{luo2025tactileroboticsoutlook}. The vast majority of these simulators are designed for vision-based tactile sensors, such as Digit~\cite{Lambeta_2020} and GelSight~\cite{s17122762}. To achieve realistic sensor simulation, early frameworks focused heavily on high-fidelity rendering, developing specialized pipelines to model complex soft-body deformations, optical responses, and marker motion fields~\cite{si2021taximexamplebasedsimulationmodel, zhao2024fots, shen2024simulation}. More recent efforts have prioritized scalability by integrating tactile rendering and collision dynamics into GPU-accelerated physics engines~\cite{akinola2025tacsl, li2025taccel, chen2026univtac}. Other works have developed simulators that explicitly model elastic deformation to generate dense force and stress fields~\cite{wang2021elastic}. By providing a physically grounded representation of contact mechanics rather than purely optical approximations, these dense models enable algorithms to reason directly about complex material interactions.
% Furthermore, differentiable simulators have been introduced to provide analytical gradients through the contact physics and sensor rendering pipelines~\cite{pmlr-v205-xu23b, si2024difftactile}, facilitating advanced control and system calibration.

Despite these advances in simulation, the transfer of learned policies to physical hardware remains a critical bottleneck~\cite{luo2025tactileroboticsoutlook}. To bridge the sim-to-real gap, prior works rely on image-to-image translation and domain adaptation, utilizing generative models and neural rendering to map real tactile images into simulated domains or vice versa~\cite{zhao2024fots, church2021tactilesimtorealpolicytransfer, Zhong2025TactGenTS}. While effective, these translation methods remain highly engineered for the specific optical mechanics of vision-based sensors. Similarly, physics-based and differentiable models still heavily rely on simplifying assumptions for material deformations~\cite{wang2021elastic}, or require tedious sensor calibration based on localized physical quantities~\cite{qi2023general, yang2024anyrotate}. In contrast, our approach sidesteps the limitations of aligning raw sensory distributions, making restrictive physical assumptions, and performing extensive hardware calibration. Instead, we reproduce identical physical stimuli across both high-fidelity simulations and real sensors, aligning these heterogeneous modalities directly within a shared embedding space.

\begin{figure*}[t!]
    \centering
    \includegraphics[width=0.99\linewidth]{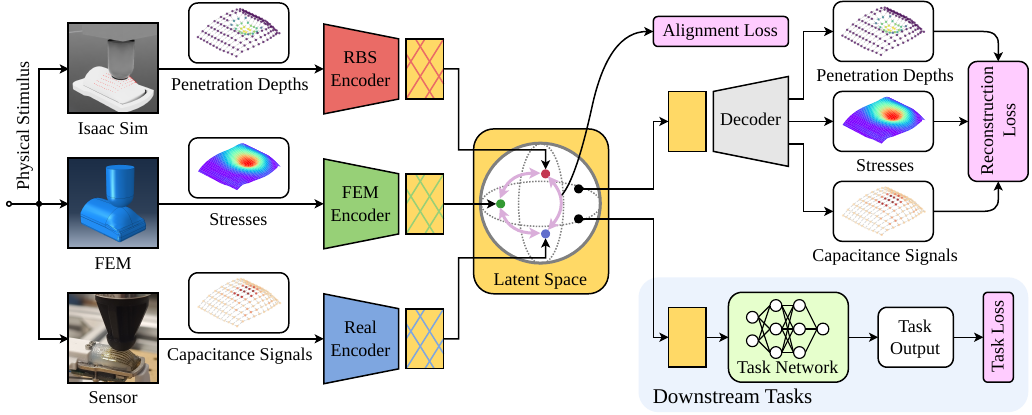}
    \caption{\textbf{Overview of the proposed multi-modal latent alignment framework}. Each modality is processed by a dedicated ViT encoder that maps observations into a shared latent space. A contrastive alignment loss encourages embeddings of the same stimulus to cluster together. A shared decoder reconstructs all modalities from any latent embedding, supervised by a reconstruction loss. The resulting representations are then used for downstream tasks, where a lightweight task network is trained on simulated embeddings and evaluated zero-shot on real sensor measurements.}
    \label{fig:tactful_overview}
    \vspace{-12pt}
\end{figure*}

\vspace{-4pt}
\subsection{Tactile Representation Learning}

Recent advancements in representation learning have significantly enhanced the integration of tactile sensing with other modalities, most notably vision and language. A common approach involves using self-supervised or contrastive learning objectives to project tactile signals into pre-trained visual or textual embedding spaces~\cite{yang2024bindingtoucheverythinglearning, dave2024multimodalvisualtactilerepresentationlearning, chi2024multi}. However, these methods focus primarily on the alignment of semantic material properties or global (scene-level) signals. They often abstract away the complex, dynamic contact mechanics and interaction forces that are essential for low-level robotic manipulation. A parallel line of research focuses on cross-sensor alignment to learn a unified latent space or explicitly translate the raw outputs of one type of tactile sensor into those of another~\cite{yang2024bindingtoucheverythinglearning, rodriguez2025contrastive, NEURIPS2022_35489258}. While this cross-device standardization is highly beneficial for mitigating hardware fragmentation and reusing datasets, it relies on mapping between physical hardware domains, thereby leaving the sim-to-real gap unaddressed. 

More closely aligned with our objective are methods that leverage multi-modal tactile representations explicitly to solve contact-rich manipulation tasks~\cite{lygerakis2024m2curl, han2025upvital, heng2025vitacformer, CAO2024104688}. Works such as M2CURL~\cite{lygerakis2024m2curl} and ViTacFormer~\cite{heng2025vitacformer} demonstrate that explicitly aligning visual and tactile spaces can significantly accelerate policy learning and improve robustness during tasks like grasping and insertion. However, these representation methods predominantly train their latent spaces entirely within simulation. By assuming access to perfect, idealized tactile feedback during both training and deployment, these approaches effectively bypass the sim-to-real transfer problem altogether.

For tactile sim-to-real transfer, the work most closely related to ours is by Narang \etal~\cite{narang2021simtorealrobotictactilesensing}, which employs neural networks to bridge the gap between FEA simulations and real-world piezoresistive sensors. They achieve this by learning latent projections that map simulated electrical signals directly to real-world sensor outputs. However, their architecture functions primarily as an asymmetric translation mechanism, explicitly projecting one modality onto another rather than extracting a shared underlying structure. In contrast, our framework leverages self- and cross-reconstruction objectives alongside contrastive alignment to project both rigid-body kinematics and high-fidelity FEA stress fields into a shared, modality-invariant latent space. This ensures that fundamentally richer contact phenomena are encoded, which we demonstrate to be highly effective for robust transfer across a diverse set of downstream manipulation tasks.

\section{METHOD}
%\mayank{We consider three sensing modalities: a real-world capacitance array, FEM-based taxel stress, and ray-casting-based penetration depth. The capacitance sensor~\cite{johannes_tactful} consists of a silicone membrane that induces complex, difficult-to-model deformations and hysteresis. This leads to the following question: How can models trained in simulation be transferred to such sensors when only limited real-world data is available? }

We propose a representation learning framework that maps heterogeneous tactile sensing modalities into a shared latent space, enabling sim-to-real transfer without requiring raw signal matching between simulation and hardware. The framework consists of two stages: (1) multi-modal latent alignment, which learns a modality-invariant embedding space via contrastive and reconstructive objectives, and (2) downstream evaluation, which probes the expressiveness and transfer capability of the learned representations by training on simulation-only data and testing on real sensor data. An overview of the full architecture is shown in~\cref{fig:tactful_overview}.

\subsection{Problem Formulation}

% \mayank{Might be nicer to ground this formulation with the capacitance-based tactile sensor example.  It is a nice example to concretize/ground the discussions around for the readers?} We decided that it is nicer to keep the implementation details away from our method, because we wanted to portray it as a method that can be used for any tactile sensor.
A central challenge in tactile sim-to-real transfer is that simulated and real sensor signals differ fundamentally in structure, noise characteristics, and physical interpretation. Rather than attempting to close this gap at the signal level, we learn a shared latent space, which we refer to as \emph{TactSpace}, in which observations from different modalities, whether real sensor measurements or simulated representations, that correspond to the same underlying physical stimulus are embedded consistently.

To this end, we assume access to a set of K sensing modalities ${M} = \{M_k\}_{k=1}^{K}$, each representing a distinct observation representation for tactile sensing. For any given physical stimulus, each modality produces a corresponding observation, and observations sharing the same stimulus are considered \emph{aligned}. The objective is to learn encoders that map these heterogeneous observations into a common embedding space, such that observations are represented similarly regardless of their source modality.

\subsection{Multi-Modal Latent Alignment}

\noindent\textbf{Architecture.}
Each input modality is processed by a dedicated Vision Transformer (ViT)~\cite{dosovitskiy2021imageworth16x16words} encoder that maps observations into a shared latent space ${Z}$. The self-attention mechanism helps model global spatial structure, which is well-suited for grid-structured tactile observations where long-range correlations between sensing elements convey contact information.
% ViT encoders are chosen for their ability to capture global spatial structure via self-attention, which is well-suited to grid-structured tactile observations where long-range correlations between sensing elements carry meaningful contact information.

% \mayank{Do you mean shared backbone but different heads? Right now it sounds one decoder does for all modalities...}
A multi-layer perceptron (MLP) decoder with a shared backbone and modality-specific heads reconstructs all modalities from a latent embedding. In this cross-reconstruction design, an embedding produced by one modality's encoder must recover the observations of all modalities. This acts as a strong regularizer. It discourages modality-specific artifacts and encourages representations that capture the underlying physical interaction in a modality-agnostic form.
% A single shared multi-layer perceptron (MLP) decoder receives any latent embedding and reconstructs all modalities simultaneously. This \emph{cross-reconstruction} design -- where the embedding produced by one modality's encoder must be sufficient to recover the observations of all other modalities -- is a key architectural choice. It acts as a strong regularizer, preventing the latent space from retaining modality-specific artifacts and instead forcing it to encode the underlying physical interaction in a modality-agnostic form.

\noindent\textbf{Training Objectives.}
The framework is trained jointly using two complementary objectives:

\noindent\underline{Alignment Loss.} A pairwise InfoNCE contrastive loss~\cite{oord2019representationlearningcontrastivepredictive} is applied across all pairs of modalities in the latent space. For each batch, embeddings from different modalities corresponding to the same physical stimulus are attracted together, while other embeddings from distinct stimuli are pushed apart. This encourages the latent space to organize according to the structure of the stimuli rather than the characteristics of individual modalities.

\noindent\underline{Reconstruction Loss.} A mean-squared error loss is applied to all cross-modal reconstructions, ensuring that the latent embeddings retain sufficient physical information to recover any modality from any other. Together with the alignment loss, this prevents representational collapse and grounds the shared embedding space in the observable signal content of each modality.

The overall training objective is a weighted combination of these two terms $L = \lambda_{\text{align}}\, L_{\text{align}} + \lambda_{\text{recon}}\, L_{\text{recon}}$,
where $\lambda_{\text{align}}$ and $\lambda_{\text{recon}}$ balance the relative contribution of each objective.

\subsection{Downstream Evaluation}

To assess the quality of the learned representations, we define a set of downstream tasks spanning classification and regression over various contact stimuli. In each case, a lightweight MLP is trained on top of frozen encoder embeddings using simulated data only and evaluated directly on real sensor measurements without any fine-tuning. This zero-shot transfer protocol tests whether the alignment has successfully abstracted away modality-specific artifacts, yielding representations that bridge the sim-to-real gap.

\subsection{Scalable Tactile Simulation}
\label{sec:tactile-sim}

We present a custom plugin for NVIDIA Isaac Lab that enables scalable tactile simulation for robot learning. Building upon the penalty-based formulation by Xu~\etal~\cite{pmlr-v205-xu23b}, the plugin models contact geometry and pressure distributions across tactile sensor surfaces via GPU-accelerated raycasting. All computations are implemented using NVIDIA Warp to run natively on the GPU. This allows the tactile simulation to be massively parallelized across thousands of concurrent environments, as shown in~\cref{fig:warp-sim}. The simulation framework is therefore well-suited for large-scale reinforcement learning and synthetic data generation pipelines. Importantly, this parallelization does not compromise physical fidelity. However, the generated tactile data may not perfectly match real-world measurements due to unmodelled mechanical deformation and transduction effects.

\begin{figure}
    \centering
    \begin{subfigure}[b]{0.31\linewidth}
        \centering
        \includegraphics[height=0.95in]{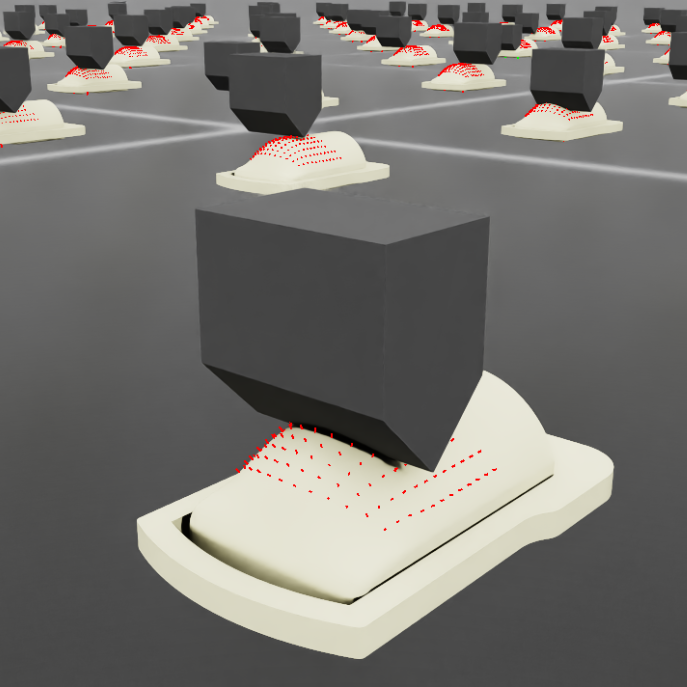}
        \caption{}
        \label{fig:warp-sim-scale}
    \end{subfigure}
    \hfill
    \begin{subfigure}[b]{0.31\linewidth}
        \centering
        \includegraphics[height=0.95in]{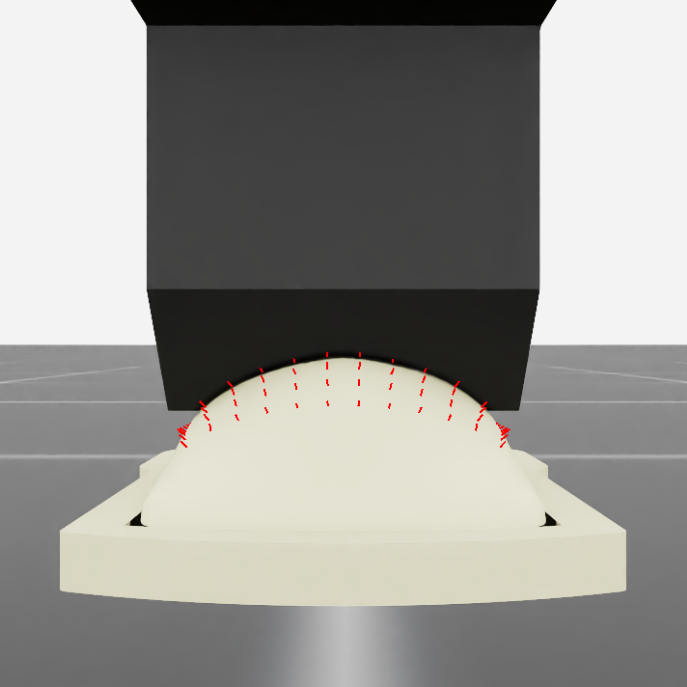}
        \caption{}
        \label{fig:warp-sim-env}
    \end{subfigure}
    \hfill
    \begin{subfigure}[b]{0.31\linewidth}
        \centering
        \includegraphics[height=0.95in]{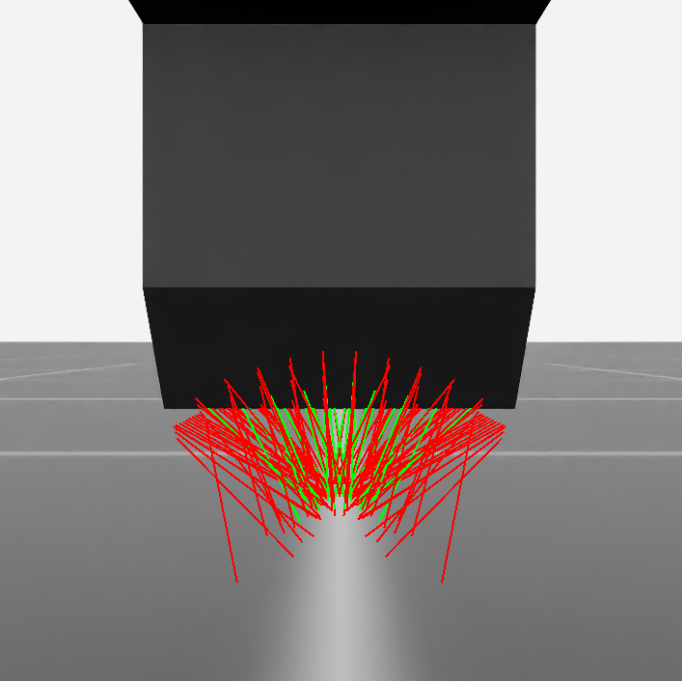}
        \caption{}
        \label{fig:warp-sim-rays}
    \end{subfigure}
    \caption{ \textbf{GPU-accelerated tactile simulation in NVIDIA Isaac Lab.} (a) Massively parallelized tactile simulation running across hundreds of concurrent environments. (b) A simulated tactile sensor in contact with a rigid object. (c) Tactile sensor with the sensor mesh hidden, exposing the underlying raycasting rays. The red rays indicate sampling directions and green rays indicate collision of the ray with a rigid object.}
    \label{fig:warp-sim}
    \vspace{-14pt}
\end{figure}

\section{EXPERIMENTAL SETUP}

This section presents the experimental evaluation of the proposed framework. We first describe the sensor, simulation environments, and dataset used for training and evaluation.

% \begin{figure}
%     \centering
%     \begin{subfigure}[b]{0.48\linewidth}
%         \centering
%         \includegraphics[width=\textwidth]{img/exp-setup-cropped-1.jpg}
%         % \caption{Caption}
%         \label{fig:exp_setup_1}
%     \end{subfigure}
%     \hfill
%     \begin{subfigure}[b]{0.48\linewidth}
%         \centering
%         \includegraphics[width=\textwidth]{img/exp-setup-cropped-2.jpg}
%         % \caption{Caption}
%         \label{fig:exp_setup_2}
%     \end{subfigure}
%     \caption{Experimental setup (rotates on its own)}
% \end{figure}

\subsection{Data Collection} 

We define a \emph{physical stimulus} as a probing interaction parameterized by:
\begin{itemize}
    \item Indenter geometry: the shape (e.g. curved, cylindrical, square) and size of the indenter;
    \item Probing location: the $(x,y)$ position of the indenter relative to the sensor surface,
    \item Probing displacement: the displacement of the indenter along the surface normal ($z$-axis) relative to the undeformed sensor surface.
\end{itemize}
A physical stimulus is uniquely specified by a particular combination of these parameters and serves as a common reference for generating paired interactions across sensing modalities. Using this formulation, we collect dense tactile data from three sources: real-world capacitance measurements (Real), rigid body simulation (RBS) using Isaac Lab~\cite{mittal2025isaaclab}, and finite element method (FEM) simulations using ABAQUS~\cite{0b112d0e5eba4b7f9768cfe1d818872e}.

\begin{figure}
    \centering
    % \begin{subfigure}[b]{0.20\linewidth}
    %     \centering
    %     \includegraphics[width=0.7in,angle=-90]{img/exp-setup-cropped-2.jpg}
    %     \caption{Probing Setup}
    %     \label{fig:probing_setup}
    % \end{subfigure}
    % \hfill
    \begin{subfigure}[b]{0.40\linewidth}
        \centering
        \includegraphics[height=0.95in]{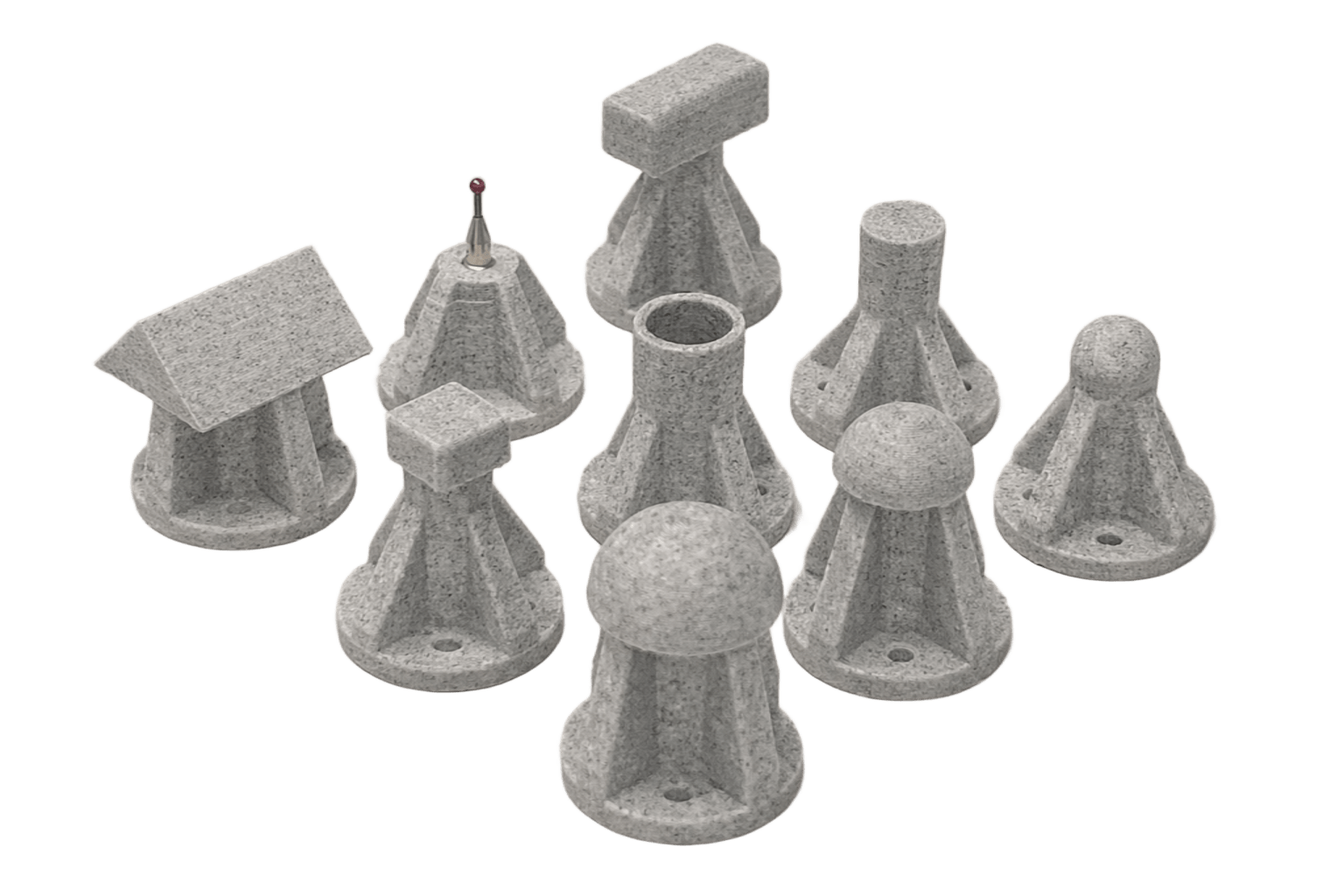}
        \caption{Alignment Data}
        \label{fig:alignment_indenters}
    \end{subfigure}
    \hfill
    \begin{subfigure}[b]{0.25\linewidth}
        \centering
        \includegraphics[height=0.95in]{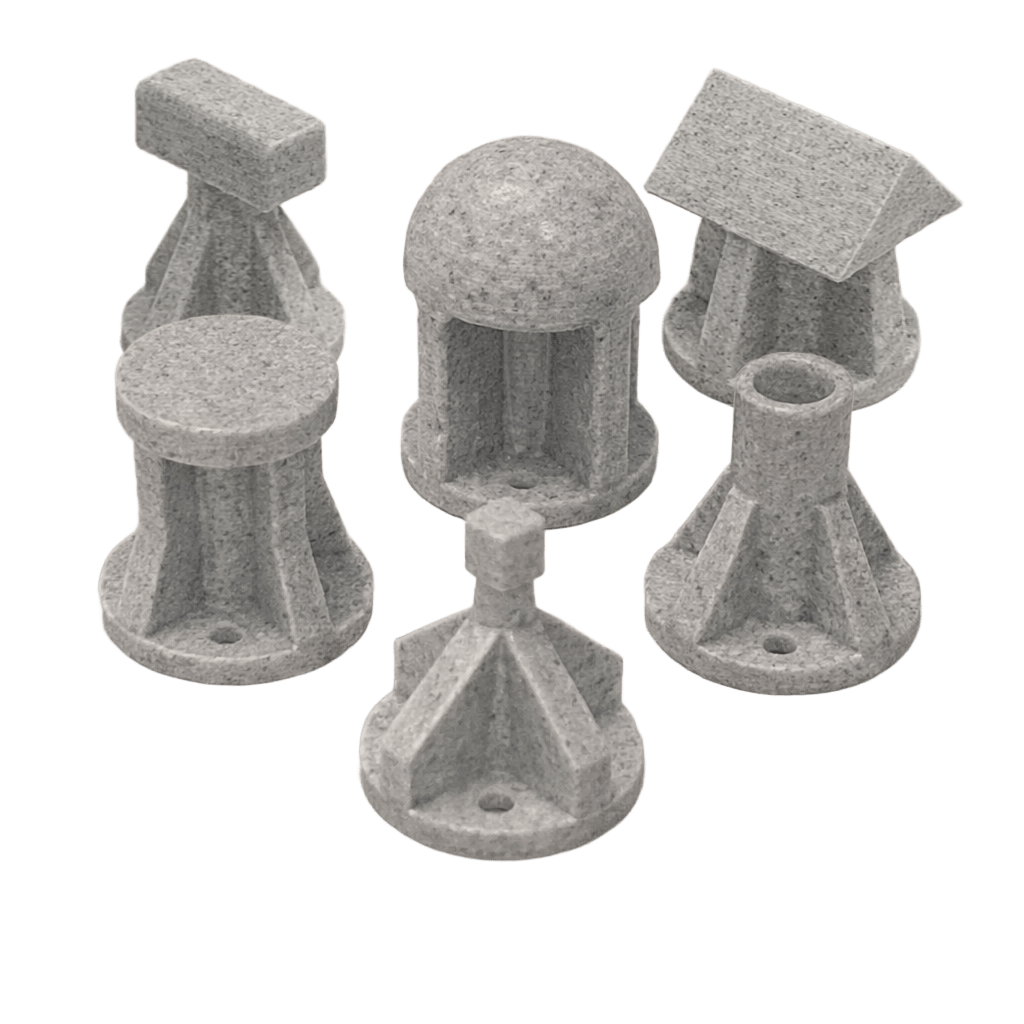}
        \caption{OOD-Size}
        \label{fig:ood_size_indernters}
    \end{subfigure}
    \hfill
    \begin{subfigure}[b]{0.25\linewidth}
        \centering
        \includegraphics[height=0.95in]{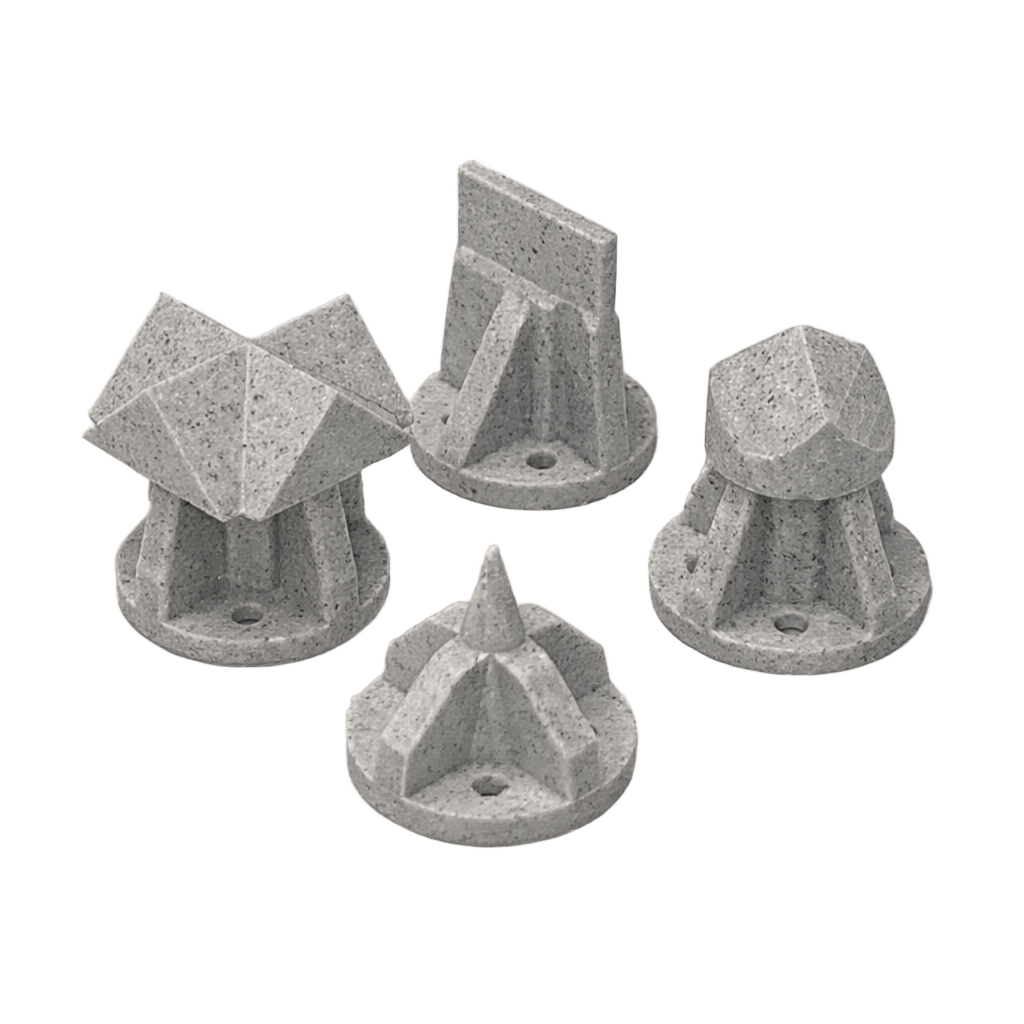}
        \caption{OOD-Shape}
        \label{fig:ood_shape_indernters}
    \end{subfigure}
    \caption{\textbf{Indenter geometries used in this work.} Set (a) is used during latent alignment, while sets (b) and (c) are reserved for downstream tasks.}
    \vspace{-14pt}
\end{figure}

\underline{Real.} We use an artificial silicone fingertip~\cite{johannes_tactful} equipped with $n_{\text{taxel}} \times n_{\text{taxel}}$ array of capacitive tactile pixels (\emph{taxels}). Each taxel produces a signal in response to local surface deformation, resulting in an image of capacitive measurements. In our sensor setup, $n_{\text{taxel}} = 12$.

The hysteresis in the sensor’s silicone layer introduces artifacts into the capacitive measurements. To address this, we incorporate temporal history by maintaining a short buffer of recent frames. This buffer helps to capture the evolution of the contact state over time. We refer to it as the \emph{capacitance modality}, $M_c \in \mathbb{R}^{n_{\text{taxel}} \times n_{\text{taxel}} \times  n_\text{hist}}$, with $n_\text{hist} = 10$.

To generate the paired real-world dataset, a repurposed 5-axis CNC milling machine executes the probing interactions according to the exact parameters defining each physical stimulus. A 6-DOF force-torque (F/T) sensor, mounted in-line with the indenter, measures the precise net contact force applied to the sensor surface. For every physical stimulus, we synchronously log the positional coordinates from the CNC machine, the ground-truth force data from the F/T sensor, and the corresponding raw capacitance measurements ($M_c$).

\underline{RBS.} Using the tactile simulation described in~\cref{sec:tactile-sim}, we treat the spatial locations of all taxels on the sensor surface as raycasting origins. Casting rays along each taxel's surface normal allows us to measure the indenter penetration depth from compliant contacts, producing the \emph{penetration depth modality} $M_p \in \mathbb{R}_{\geq 0}^{n_{\text{taxel}} \times n_{\text{taxel}}}$.

\underline{FEM.} We use ABAQUS to simulate the probing process on a finite element model of the tactile sensor. The model tracks a stress field for different load cases, yielding the \emph{taxel stress modality} $M_s \in \mathbb{R}^{n_{\text{taxel}} \times n_{\text{taxel}} \times 6}$. The FEM simulation additionally provides access to quantities not directly observable on the physical sensor, including indenter reaction forces, which are subsequently used as ground-truth labels for downstream force prediction tasks.

\subsection{Alignment and Downstream Data Splits}

The dataset is organized into two non-overlapping sets based on the training phase. The alignment dataset, $\mathbf{D}_A=~\mathbf{D}_A^\text{train}~\cup~\mathbf{D}_A^\text{test}$, consists of the indenter shapes used to train the multi-modal encoders, while the downstream dataset $\mathbf{D}_D=~\mathbf{D}_D^\text{train}~\cup~\mathbf{D}_D^\text{test}$, is reserved for training and evaluating task performance.
% The indenter geometries used to collect the alignment data are shown in Figure~\ref{fig:alignment_indenters}.
%
\cref{fig:alignment_indenters} shows the nine indenter geometries used for alignment.
For downstream tasks on $\mathbf{D}_D$, we report performance on three test partitions of increasing difficulty. The in-distribution holdout contains the same geometries and sizes as the alignment data but uses held-out locations. The OOD-Size partition uses six indenters with the same shapes but with previously unseen sizes and orientations (\cref{fig:ood_size_indernters}). The OOD-Shape partition contains four entirely new indenter geometries (\cref{fig:ood_shape_indernters}).

% For downstream evaluation on $\mathbf{D}_D$, we report performance across three test splits of increasing difficulty: an \emph{in-distribution holdout} (same geometries and sizes as latent alignment, held-out locations), \emph{OOD-Size} (same shapes, unseen sizes), shown in~\cref{fig:ood_size_indernters}, and \emph{OOD-Shape} (entirely unseen indenter geometries), in~\cref{fig:ood_shape_indernters}. Task networks are trained exclusively on simulated data and evaluated zero-shot on real capacitance measurements throughout, with no access to real sensor data during training.

Data is collected for each of the three modalities (Real, RBS, FEM) across 840 probing trajectories, yielding 32,942 samples for $M_c$ and $M_p$ each and 26,107 samples for $M_d$. The physical stimuli span 19 indenters across 12 distinct shapes. Penetration depth has a median of 62.63~$\mu$m (range: 0.00--1330.98~$\mu$m), and indenter force has a median of 498.66~mN (range: 1.02--6837.94~mN).
The 840 trajectories are partitioned into three subsets: 462 in-distribution, 210 OOD-Size, and 168 OOD-Shape. The in-distribution trajectories are further divided equally, with 231 trajectories allocated to the alignment dataset $\mathbf{D}_A$ and 231 to the in-distribution holdout for the downstream dataset $\mathbf{D}_D$. The OOD-Size and OOD-Shape trajectories are assigned exclusively to $\mathbf{D}_D$.
Within each subset, trajectories are further divided into training and test sets using a 75\%–25\% split based on probing location.
% Each subset is split 75\%--25\% into training and test sets based on probing location.

% \begin{table}[h]
%     \centering
%     \caption{Dataset Statistics}
%     \label{tab:dataset_stats}
%     \begin{tabular}{lccc}
%         \toprule
%         \textbf{Property} & \textbf{Real} & \textbf{FEM} & \textbf{Sim} \\
%         \midrule
%         No. of probing trajectories & 840 & 840 & 840 \\
%         No. of samples & 32942 & 26107 & 32942 \\
%         Total number of different indenters & \multicolumn{3}{c}{19} \\
%         Distinct indenter shapes & \multicolumn{3}{c}{12} \\
%         Penetration depth (median, min, max) [$\mu$m] & \multicolumn{3}{c}{62.63 (0.00, 1330.98)} \\
%         Indenter force (median, min, max) [mN] & \multicolumn{3}{c}{498.66 (1.02, 6837.94)} \\
%         \bottomrule
%     \end{tabular}
% \end{table}
\vspace{-3pt}
\subsection{Downstream Tasks}
To assess the cross-modal transferability of the learned representation space, we define three downstream tasks:

\noindent\underline{Indenter Shape Classification.}
Given a tactile observation, the model predicts the indenter shape category via a multi-label classification objective. Performance is measured by sample-weighted classification accuracy (in \%), denoted \textbf{Acc.}\ in~\cref{sec:results}.

\noindent\underline{Indenter Shape Reconstruction.}
The model predicts per-taxel penetration depths along surface normals, providing a geometric reconstruction of the contact surface. Performance is measured by the pointwise mean absolute error over predicted penetration depths (in $\mu$m), denoted \textbf{Pen.}\ in~\cref{sec:results}.

\noindent\underline{Indenter Force Prediction.}
The model estimates the total reaction force exerted on the indenter arising from the sensor-indenter mechanical interaction. Performance is measured by mean absolute error (in mN), denoted \textbf{Force}\ in~\cref{sec:results}.

It is important to note that the task networks are trained exclusively on simulated data and evaluated zero-shot on real capacitance measurements.

\vspace{-2pt}
\section{RESULTS}
\label{sec:results}
In this section, we empirically validate our multi-modal representation-learning framework. We first analyze the cross-modal alignment of the learned latent space (\cref{sec:latent_alignment}) before demonstrating its effectiveness for zero-shot sim-to-real transfer on physical hardware (\cref{sec:zero_shot}). We then present a comprehensive ablation study detailing the trade-offs of different modality and loss configurations (\cref{sec:ablation}). Finally, we explore the specific benefits and physical limitations of scaling up highly parallelizable simulated data for representation pre-training (Section V-D).

\begin{figure}
    \centering
    % \begin{subfigure}[b]{0.3\linewidth}
    %     \centering
    %     \includegraphics[width=\textwidth]{img/reconstruction_and_force_prediction_1.png}
    %     \caption{$d = 600 \mu m$.}
    % \end{subfigure}
    % \hfill
    % \begin{subfigure}[b]{0.3\linewidth}
    %     \centering
    %     \includegraphics[width=\textwidth]{img/reconstruction_and_force_prediction_2.png}
    %     \caption{$d = 1400 \mu m$.}
    % \end{subfigure}
    % \hfill
    % \begin{subfigure}[b]{0.3\linewidth}
    %     \centering
    %     \includegraphics[width=\textwidth]{img/reconstruction_and_force_prediction_3.png}
    %     \caption{$d = 2400 \mu m$.}
    % \end{subfigure}

    \begin{subfigure}[b]{\linewidth}
        \centering
        \includegraphics[width=0.95\linewidth]{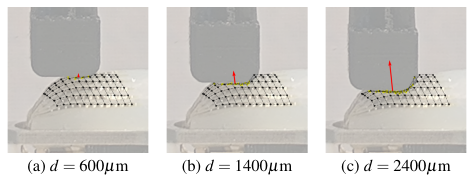}
    \end{subfigure}

    \begin{subfigure}[b]{\linewidth}
        \centering
        \includegraphics[width=0.95\linewidth]{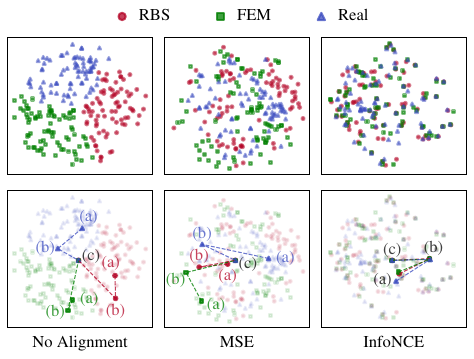}
    \end{subfigure}
        
    \caption{\textbf{Shape reconstruction and force prediction.} Indenter shape reconstruction and force prediction working in tandem. The probing depth $d$ is defined as the downwards displacement of the indenter from the point of first contact. \textbf{T-SNE visualization of latent embeddings.} Embeddings generated from different modalities with no alignment loss, mean-squared error (MSE) alignment loss, and InfoNCE alignment loss. The highlighted points in the third row correspond to the physical stimuli as shown in the first row.} 
    \label{fig:latent_alignment_trajectory}
    \vspace{-12pt}
\end{figure}

\begin{figure}[t!]
    \centering
    \includegraphics[width=0.95\linewidth]{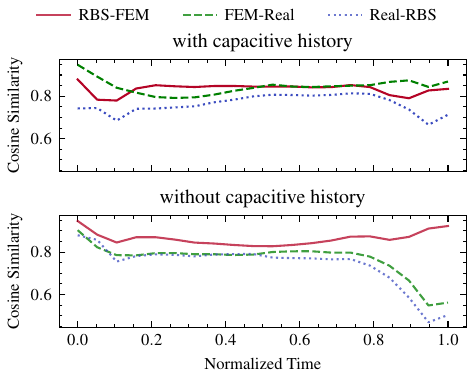}
    \caption{\textbf{Cross-modal cosine similarity.} Cross-modal cosine similarity between latent embeddings over a normalized contact trajectory, averaged across all probing experiments. Normalized time \mbox{$t \in [0, 0.5]$} corresponds to the indenter pressing into the sensor, while \mbox{$t \in [0.5, 1]$} corresponds to the indenter retracting.}
    \label{fig:cosine_similarity_trajectory}
    \vspace{-16pt}
\end{figure}

\begin{table*}[t!]
    \centering
    \caption{\label{tab:zero_shot_transfer}\textsc{\textbf{Zero-Shot Sim-to-Real Transfer On Downstream Tasks.}} Per task, a single network is trained on latents from the simulated modalities ($M_p$, $M_s$) only. On the evaluation dataset, the network is tested on latents from each modality independently. %: simulated penetration depth ($M_p$, RBS), simulated taxel stress ($M_s$, FEM), and real capacitance ($M_c$, Real).
    % Notably, despite never being exposed to real measurements during training, the network achieves comparable downstream performance when evaluated on real latents, demonstrating effective zero-shot sim-to-real transfer through the aligned representation space.
    Despite no real sensor data during downstream task training, performance on real-sensor latents is comparable, demonstrating effective zero-shot sim-to-real transfer.
    Values in parentheses denote the 20th--80th percentile range. \textbf{Bold} indicates best performance, \underline{underline} indicates second best.}
    \resizebox{1.0\textwidth}{!}{%
        \setlength{\tabcolsep}{3pt}
        \rowcolors{3}{gray!10}{white}
        \begin{tabular}{l l||c c c | c c c | c c c}
            \toprule
            \textbf{Data} & \multirow{2}{*}{\textbf{Modality}} & \multicolumn{3}{c|}{\textbf{In-Distribution Holdout}} & \multicolumn{3}{c|}{\textbf{OOD-Size}} & \multicolumn{3}{c}{\textbf{OOD-Shape}} \\
            \textbf{Origin} & & \textbf{Acc.}~[\%]$\uparrow$ & \textbf{Pen.}~[$\mu$m]$\downarrow$ & \textbf{Force}~[mN]$\downarrow$ & \textbf{Acc.}~[\%]$\uparrow$ & \textbf{Pen.}~[$\mu$m]$\downarrow$ & \textbf{Force}~[mN]$\downarrow$ & \textbf{Acc.}~[\%]$\uparrow$ & \textbf{Pen.}~[$\mu$m]$\downarrow$ & \textbf{Force}~[mN]$\downarrow$ \\
            \midrule
            RBS & Penetration Depth ($M_p$) & \textbf{64.30} & \underline{16} {\tiny (10 - 37)} & \underline{65} {\tiny (50 - 126)} & \textbf{71.75} & \underline{18} {\tiny (9 - 40)} & 67 {\tiny (50 - 125)} & \underline{44.50} & 20 {\tiny (14 - 28)} & 85 {\tiny (51 - 144)} \\
            FEM & Taxel Stress ($M_d$) & 55.96 & \textbf{12} {\tiny (5 - 25)} & \textbf{43} {\tiny (23 - 79)} & \underline{65.00} & \textbf{14} {\tiny (9 - 37)} & \textbf{51} {\tiny (26 - 85)} & 25.00 & \textbf{6} {\tiny (3 - 14)} & \textbf{27} {\tiny (20 - 51)} \\
            \midrule
            Real & Capacitance ($M_c$) & \underline{61.09} & 26 {\tiny (9 - 59)} & \underline{65} {\tiny (37 - 118)} & 51.84 & 39 {\tiny (16 - 76)} & \underline{59} {\tiny (38 - 95)} & \textbf{45.25} & \underline{13} {\tiny (6 - 41)} & \underline{63} {\tiny (36 - 112)} \\
            \bottomrule
        \end{tabular}
    }
    \vspace{-4pt}
\end{table*}

\begin{table*}[t!]
    \centering
    \caption{\label{tab:alignment_loss}\textsc{\textbf{Effect of alignment loss on downstream performance.}} InfoNCE outperforms MSE and no-alignment baselines across nearly all metrics and evaluation splits. Values in parentheses denote the 20th--80th percentile range. \textbf{Bold} indicates best performance; \underline{underline} indicates second best.}
    \rowcolors{3}{gray!10}{white}
    \resizebox{0.9\textwidth}{!}{%
        \setlength{\tabcolsep}{3pt}
        \begin{tabular}{l || c c c | c c c | c c c}
            \toprule
             \multirow{2}{*}{\textbf{Alignment Loss}} & \multicolumn{3}{c|}{\textbf{In-Distribution Holdout}} & \multicolumn{3}{c|}{\textbf{OOD-Size}} & \multicolumn{3}{c}{\textbf{OOD-Shape}} \\
             & \textbf{Acc.}~[\%]$\uparrow$ & \textbf{Pen.}~[$\mu$m]$\downarrow$ & \textbf{Force}~[mN]$\downarrow$ & \textbf{Acc.}~[\%]$\uparrow$ & \textbf{Pen.}~[$\mu$m]$\downarrow$ & \textbf{Force}~[mN]$\downarrow$ & \textbf{Acc.}~[\%]$\uparrow$ & \textbf{Pen.}~[$\mu$m]$\downarrow$ & \textbf{Force}~[mN]$\downarrow$ \\
            \midrule
            No Alignment & 36.72 & 37 {\tiny(12 - 82)} & 89 {\tiny(66 - 131)} & 39.69 & 74 {\tiny(49 - 111)} & 86 {\tiny(66 - 122)} & 26.15 & \textbf{10} {\tiny(2 - 55)} & 77 {\tiny(57 - 107)} \\
            MSE & \underline{52.50} & \underline{35} {\tiny(21 - 66)} & \underline{68} {\tiny(38 - 122)} & \underline{49.81} & \underline{46} {\tiny(19 - 84)} & \underline{64} {\tiny(38 - 113)} & \underline{44.15} & 14 {\tiny(6 - 51)} & \underline{67} {\tiny(40 - 111)} \\
            \midrule
            InfoNCE (Ours) & \textbf{61.09} & \textbf{26} {\tiny(9 - 59)} & \textbf{65} {\tiny(37 - 118)} & \textbf{51.84} & \textbf{39} {\tiny(16 - 76)} & \textbf{59} {\tiny(38 - 95)} & \textbf{45.25} & \underline{13} {\tiny(6 - 41)} & \textbf{63} {\tiny(36 - 112)} \\
            \bottomrule
        \end{tabular}
    }
    \vspace{-10pt}
\end{table*}

\subsection{Latent Space Alignment}
\label{sec:latent_alignment}
We first assess whether the contrastive alignment objective produces a well-structured shared latent space across modalities. \cref{fig:latent_alignment_trajectory} shows two-dimensional t-SNE projections~\cite{JMLR:v9:vandermaaten08a} of latent embeddings under three training configurations: no alignment loss, MSE alignment loss, and InfoNCE alignment loss. Without alignment, embeddings from different modalities form distinct clusters, indicating that the encoder outputs remain modality-specific. Introducing an MSE alignment loss reduces separation but fails to produce a coherent joint structure. In contrast, the InfoNCE objective results in a uniform intermixing of embeddings from all three modalities, confirming that the learned latent space is organized by the underlying physical stimulus rather than the observation source.

Next, we examine the role of temporal history in the capacitive data.
\cref{fig:cosine_similarity_trajectory} illustrates the evolution of pairwise cosine similarity between cross-modal embeddings along the probing trajectory. When temporal (capacitive) history is included, the cosine similarity between embeddings from FEM, RBS, and real sensors remains consistently similar throughout the trajectory. Without history, we observe a pronounced drop in pairwise cosine similarity, particularly between FEM-Real and RBS-Real embeddings, toward the latter portion of the trajectory. This degradation arises from capacitive and material hysteresis effects. Lacking temporal context, the embeddings cannot capture history-dependent behavior in the capacitance signals, leading to misalignment.

\subsection{Zero-Shot Sim-to-Real Transfer}
\label{sec:zero_shot}

Having established that the InfoNCE objective produces a well-aligned latent space, we evaluate whether this alignment translates into effective zero-shot sim-to-real transfer on downstream tasks. Task networks are trained on the embeddings from simulated modalities ($M_p$ from Isaac Sim, $M_s$ from FEM) and evaluated on real capacitance measurements ($M_c$), \emph{without any fine-tuning on real (capacitance) data}.
% We report the performance across three downstream tasks: indenter shape identification (classification accuracy), contact force prediction (force error in mN), and indenter shape reconstruction (penetration depth error in $\mu$m) and three evaluation datasets.

% \vspace{-5pt}

As shown in Table~\ref{tab:zero_shot_transfer}, our aligned latent space, or \emph{TactSpace}, enables highly effective zero-shot sim-to-real transfer across all evaluation splits, robustly handling even out-of-distribution indentations. Although the task networks are trained exclusively on idealized and noiseless simulated data, their performance remains comparable when evaluated directly on real capacitance embeddings. This confirms that the latent alignment allows the model interpret raw hardware measurements despite never seeing them during training.
Furthermore, the source modalities exhibit complementary strengths. Rigid-body representations ($M_p$) excel at geometric predictions, while finite-element representations ($M_s$) dominate force estimation.
This validates that our framework is able to capture modality-agnostic contact physics.

Table~\ref{tab:alignment_loss} further ablates the choice of alignment loss on downstream zero-shot transfer performance. Across all evaluation splits and metrics, Task networks trained on InfoNCE-aligned latent spaces consistently outperform those trained on MSE-aligned or unaligned spaces. These results support the qualitative findings in~\cref{sec:latent_alignment} and confirm that the quality of cross-modal alignment directly affects sim-to-real transfer of the learned representations.

% TODO: This is wrong

% Across all tasks and splits, sim-trained encoders achieve performance competitive with the real encoder upper bound, demonstrating that the aligned latent space supports zero-shot transfer without any access to real sensor data. The Isaac Sim encoder ($M_p$) consistently outperforms the FEM encoder ($M_s$) on shape-related tasks, while the FEM encoder yields lower force prediction error, reflecting the complementary physical content of the two simulation modalities.

% \textcolor{red}{TODO: Large Scale Sim Data row: results pending. This row will demonstrate convergence of sim-trained task networks toward real-encoder performance as the volume of simulated training data increases, highlighting the scalability advantage of the proposed framework.}

\subsection{Representation Expressiveness and Modality Ablation}
\label{sec:ablation}

We now aim to answer two key questions: (1) how does the choice of simulated modality affect downstream task performance, and (2) to what extent does the cross-reconstruction objective improve the learned representation?
Table~\ref{tab:generalization} presents a comprehensive ablation over pre-training data sources, reconstruction loss configurations, and task training modalities. We also evaluate our zero-shot approach against a supervised baseline trained directly on real sensor measurements. Since this network predicts directly from physical sensor measurements, it completely bypasses the sim-to-real gap and serves as a practical upper bound for performance.

\begin{table*}[t]
    \centering
    % \caption{\label{tab:generalization}\textsc{\textbf{Generalization.}} Performance of models trained on different data and loss configurations, evaluated on in-distribution and out-of-distribution test sets. \textbf{Alignment Data} and \textbf{Task Data} indicate which data sources are used during pre-training and downstream task training, respectively. Reconstruction Loss (\textbf{Recon. Loss}) denotes which signal modalities are used during pre-training: capacitance (\textbf{Cap.}), penetration depth (\textbf{Pen.}), and stress (\textbf{Str.}). Results are reported across three evaluation splits: the \textit{in-distribution holdout} set, the \textit{OOD-Size} set, and the \textit{OOD-Shape} set. Values in parentheses denote the 20th–80th percentile range, \textbf{bold} indicates best performance, \underline{underline} indicates second best.}
    \caption{\label{tab:generalization}\textsc{\textbf{Representation Expressiveness and Modality Ablation.}} Performance of models trained on different data and loss configurations, evaluated on in-distribution and out-of-distribution test sets. \textbf{Alignment Data} and \textbf{Task Data} indicate which data sources are used during latent alignment and downstream task training, respectively. Reconstruction Loss (\textbf{Recon. Loss}) denotes which signal modalities are reconstructed during latent alignment: capacitance ($M_c$), penetration depth ($M_p$), and stress ($M_s$). Values in parentheses denote the 20th–80th percentile range, \textbf{bold} indicates best performance, \underline{underline} indicates second best. The last row (in yellow) is a baseline trained only on real sensor data and serves as an upper bound on performance.}
    \rowcolors{3}{gray!10}{white}
    \resizebox{1.0\textwidth}{!}{%
        \setlength{\tabcolsep}{3pt}
        \begin{tabular}{cccc|ccc|ccc||ccc|ccc|ccc}
            \toprule
            \multicolumn{4}{c|}{\textbf{Alignment Data}} & \multicolumn{3}{c|}{\textbf{Recon. Loss}} & \multicolumn{3}{c||}{\textbf{Task Data}} & \multicolumn{3}{c|}{\textbf{In-Distribution Holdout}} & \multicolumn{3}{c|}{\textbf{OOD-Size}} & \multicolumn{3}{c}{\textbf{OOD-Shape}}   \\
            % \cmidrule(lr){11-13} \cmidrule(lr){14-16} \cmidrule(lr){17-19} 
            $M_c$ & $M_p$ & $M_s$ & \textbf{Split} & $M_c$ & $M_p$ & $M_s$ & $M_c$ & $M_p$ & $M_s$ & \textbf{Acc.} [\%]$\uparrow$ & \textbf{Pen.} [$\mu$m]$\downarrow$ & \textbf{Force} [mN]$\downarrow$ & \textbf{Acc.} [\%]$\uparrow$ & \textbf{Pen.} [$\mu$m]$\downarrow$ & \textbf{Force} [mN]$\downarrow$ & \textbf{Acc.} [\%]$\uparrow$ & \textbf{Pen.} [$\mu$m]$\downarrow$ & \textbf{Force} [mN]$\downarrow$   \\
            \midrule
            \checkmark & \checkmark & \checkmark & $\mathbf{D}_A^\text{train}$ & -- & -- & -- & -- & \checkmark & \checkmark & \underline{56.35} & \underline{45} {\tiny (7 - 180)} & \textbf{60} {\tiny (34 - 116)} & 42.58 & 79 {\tiny (34 - 206)} & \underline{61} {\tiny (38 - 104)} & 41.70 & \textbf{7} {\tiny (1 - 50)} & 65 {\tiny (38 - 123)} \\
            \checkmark & \checkmark & -- & $\mathbf{D}_A^\text{train}$ & \checkmark & \checkmark & -- & -- & \checkmark & -- & 55.49 & 58 {\tiny (11 - 115)} & 78 {\tiny (55 - 149)} & \underline{47.44} & 68 {\tiny (15 - 113)} & 75 {\tiny (54 - 147)} & \underline{42.80} & 13 {\tiny (2 - 63)} & 74 {\tiny (52 - 129)} \\
            \checkmark & -- & \checkmark & $\mathbf{D}_A^\text{train}$ & \checkmark & -- & \checkmark & -- & -- & \checkmark & 54.94 & 48 {\tiny (8 - 186)} & \underline{63} {\tiny (37 - 103)} & 45.05 & \underline{66} {\tiny (13 - 186)} & 63 {\tiny (38 - 95)} & 29.10 & \underline{8} {\tiny (2 - 53)} & \textbf{51} {\tiny (32 - 96)} \\
            \midrule
            \checkmark & \checkmark & \checkmark & $\mathbf{D}_A^\text{train}$ & \checkmark & \checkmark & \checkmark & -- & \checkmark & \checkmark & \textbf{61.09} & \textbf{26} {\tiny (9 - 59)} & 65 {\tiny (37 - 118)} & \textbf{51.84} & \textbf{39} {\tiny (16 - 76)} & \textbf{59} {\tiny (38 - 95)} & \textbf{45.25} & 13 {\tiny (6 - 41)} & \underline{63} {\tiny (36 - 112)} \\
            \midrule
            \midrule
            \checkmark & \checkmark & \checkmark & $\mathbf{D}_A^\text{train} \cup \mathbf{D}_D^\text{train}$ & \checkmark & \checkmark & \checkmark & -- & \checkmark & \checkmark & 61.91 & 25 {\tiny (6 - 60)} & 62 {\tiny (41 - 109)} & 70.66 & 49 {\tiny (29 - 85)} & 67 {\tiny (44 - 124)} & 56.05 & 8 {\tiny (4 - 38)} & 72 {\tiny (36 - 130)} \\
            \rowcolor{yellow!30}
            \checkmark & -- & -- & $\mathbf{D}_A^\text{train}$ & \checkmark & -- & -- & \checkmark & -- & -- & 76.10 & 18 {\tiny (8 - 51)} & 20 {\tiny (10 - 46)} & 83.03 & 16 {\tiny (7 - 36)} & 18 {\tiny (10 - 34)} & 69.75 & 3 {\tiny (1 - 18)} & 19 {\tiny (9 - 46)} \\
            \bottomrule
        \end{tabular}
    }
\end{table*}

\begin{table*}[t!]
    \centering
    \caption{\label{tab:large_scale_sim}\textsc{\textbf{Scaling up with simulated data.}} We evaluate the effect of augmenting training with large-scale simulated data across three downstream tasks and evaluation splits. Training configurations vary in the use of real capacitance data ($M_c$), physics-based penetration depths ($M_p$), and FEM-derived stress fields ($M_s$). Values in parentheses denote the 20th--80th percentile range. \textbf{Bold} indicates best performance; \underline{underline} indicates second best. The last row (in yellow) is a baseline trained only on real sensor data.}
    \rowcolors{3}{gray!10}{white}
    \resizebox{0.9\textwidth}{!}{%
        \setlength{\tabcolsep}{3pt}
        \begin{tabular}{l | c c c || c c c | c c c | c c c }
            \toprule
             \multicolumn{4}{c||}{\textbf{Training Data}} & \multicolumn{3}{c|}{\textbf{In-Distribution Holdout}} & \multicolumn{3}{c|}{\textbf{OOD-Size}} & \multicolumn{3}{c}{\textbf{OOD-Shape}} \\
             \multicolumn{1}{c}{\textbf{Configurations}} & $M_c$ & $M_p$ & $M_s$ & \textbf{Acc.}~[\%]$\uparrow$ & \textbf{Pen.}~[$\mu$m]$\downarrow$ & \textbf{Force}~[mN]$\downarrow$ & \textbf{Acc.}~[\%]$\uparrow$ & \textbf{Pen.}~[$\mu$m]$\downarrow$ & \textbf{Force}~[mN]$\downarrow$ & \textbf{Acc.}~[\%]$\uparrow$ & \textbf{Pen.}~[$\mu$m]$\downarrow$ & \textbf{Force}~[mN]$\downarrow$ \\
            \midrule
            RBS + FEM & -- & \checkmark & \checkmark & 61.09 & \textbf{26} {\tiny (9 - 59)} & \underline{65} {\tiny (37 - 118)} & 51.84 & \underline{39} {\tiny (16 - 76)} & \textbf{59} {\tiny (38 - 95)} & 45.25 & 13 {\tiny (6 - 41)} & \underline{63} {\tiny (36 - 112)} \\
            RBS (1x) & -- & \checkmark & -- & 58.78 & 34 {\tiny (7 - 101)} & 76 {\tiny (54 - 146)} & 49.41 & 80 {\tiny (37 - 161)} & 77 {\tiny (56 - 131)} & 38.90 & 10 {\tiny (2 - 68)} & 72 {\tiny (52 - 129)} \\
            RBS (5x) & -- & \checkmark & -- & 65.32 & 38 {\tiny (13 - 109)} & 75 {\tiny (57 - 147)} & 58.84 & 48 {\tiny (12 - 121)} & 79 {\tiny (56 - 139)} & 40.50 & \textbf{8} {\tiny (1 - 59)} & 73 {\tiny (52 - 128)} \\
            RBS (15x) & -- & \checkmark & -- & \underline{70.81} & 34 {\tiny (8 - 94)} & 78 {\tiny (59 - 146)} & \textbf{64.33} & 47 {\tiny (23 - 88)} & 81 {\tiny (63 - 140)} & \textbf{46.80} & \textbf{8} {\tiny (1 - 47)} & 75 {\tiny (55 - 128)} \\
            RBS (30x) & -- & \checkmark & -- & \textbf{70.85} & 33 {\tiny (10 - 84)} & 77 {\tiny (61 - 139)} & \underline{63.37} & 49 {\tiny (12 - 102)} & 80 {\tiny (64 - 138)} & \underline{45.45} & 9 {\tiny (1 - 50)} & 73 {\tiny (57 - 130)} \\
            RBS (30x) + FEM & \multirow{1}{*}{--} & \multirow{1}{*}{\checkmark} & \multirow{1}{*}{\checkmark} & \multirow{1}{*}{69.00} & \multirow{1}{*}{\underline{28} {\tiny (10 - 74)}} & \multirow{1}{*}{\textbf{59} {\tiny (35 - 96)}} & \multirow{1}{*}{62.99} & \multirow{1}{*}{\textbf{37} {\tiny (12 - 87)}} & \multirow{1}{*}{\textbf{59} {\tiny (37 - 91)}} & \multirow{1}{*}{41.05} & \multirow{1}{*}{\textbf{8} {\tiny (3 - 39)}} & \multirow{1}{*}{\textbf{49} {\tiny (30 - 87)}} \\ \midrule \midrule
            \rowcolor{yellow!30}
            Real & \checkmark & -- & -- & 76.10 & 18 {\tiny (8 - 51)} & 20 {\tiny (10 - 46)} & 83.03 & 16 {\tiny (7 - 36)} & 18 {\tiny (10 - 34)} & 69.75 & 3 {\tiny (1 - 18)} & 19 {\tiny (9 - 46)} \\
            \bottomrule
        \end{tabular}
    }
    \vspace{-8pt}
\end{table*}

As expected, the real-data baseline (trained directly on capacitance measurements) achieves the highest overall performance. Our proposed latent representation model, however, delivers highly competitive results. The small gap in force prediction arises because the our models are trained on FEM approximations of the net force, whereas the baseline uses the ground-truth force-torque measurements, which are also used as evaluation labels. Despite this structural advantage, collecting labeled real-world data, especially on-policy, is prohibitively slow and requires specialized hardware, making it impractical for standard robot learning workflows.

Among the latent representation models, the model pre-trained with all three modalities and full cross-reconstruction supervision achieves the best or near-best performance across all tasks, confirming that each modality contributes complementary physical information to the shared latent space. While joint training can introduce a slight performance penalty on any single task compared to the best specialist configuration, this reflects an inherent generalist--specialist tradeoff: the jointly trained encoder produces a richer, more broadly capable representation at a modest cost of task-specific optimality. 

Several trends are observable consistently across evaluation splits. Penetration-depth supervision is particularly beneficial for geometric tasks such as shape reconstruction and indenter classification, while FEM-based stress data provides strong signals for force prediction, consistent with the physical quantities each simulation captures. This suggests that if the target task is known during deployment, a specialist encoder trained on the most task-relevant simulation source can yield peak performance on that specific task. However, such specialist models come at the cost of generality, resulting in poor performance on other tasks because their latent embeddings lack complementary physical information. 

Crucially, our ablations validate the necessity of the joint training objective. Removing the reconstruction loss entirely while retaining the contrastive alignment objective results in a measurable drop in performance across downstream tasks. This indicates that the reconstruction objective is important for preventing representational collapse and preserving task-relevant information across modalities.

% Conversely, omitting the alignment objective yields disjoint modality embeddings, severely degrading zero-shot sim-to-real transfer and overall task performance. % TODO

Finally, we also report results for embeddings aligned on the entire dataset ($\mathbf{D}_A^\text{train} \cup \mathbf{D}_D^\text{train}$), \ie including out-of-distribution (OOD) indenters. Performance across OOD splits improves significantly when these OOD data are included in the latent alignment phase, demonstrating that the learned representation benefits greatly from data diversity.

\subsection{Scaling Up with Simulated Data}
\label{sec:scaling_up_image}

With the ability to scale data generation using simulation, we investigate its impact on downstream task performance. Table~\ref{tab:large_scale_sim} shows the effects of increasing the volume of the geometric RBS data. We report the zero-shot transfer performance of downstream networks trained on progressively increasing number of simulated samples. We generate data at more probing locations for the given set of indenters.
% \mayank{Needs description on how scaling up is done. More probing locations per alignement indenters? Or more downstream indenters?}
We observe a clear positive trend for geometric tasks: indenter classification and penetration depth estimation improve noticeably, approaching the real-data upper bound on the in-distribution set.
% Table~\ref{tab:large_scale_sim} illustrates the effects of scaling the volume of simulated geometric data on downstream task performance. We observe a clear positive trend in geometric tasks, as the size of the simulated dataset increases. Indenter classification and penetration depth estimation improve notably, approaching the real-data upper bound on the in-distribution set.
While scaling simulated data yields stronger performance on OOD splits, the performance gap to the real upper bound is noticeably wider than on the in-distribution set. This indicates that while massive simulation data improves the overall task performance, an underlying alignment gap persists when transferring to entirely novel shapes.
Furthermore, while data scaling significantly enhances geometric tasks, it provides negligible benefits for force prediction. In contrast, incorporating complementary multi-physics modalities, such as simulated stress, provides substantial gains for force metrics, outperforming the benefits of simply increasing the volume of single-modality data.

\section{CONCLUSION}

We present a multi-modal representation learning framework for tactile sim-to-real transfer that aligns heterogeneous sensing modalities within a shared latent space. By combining a cross-reconstruction objective with InfoNCE contrastive alignment, our approach learns modality-invariant embeddings that capture the physical structure of contact interactions without requiring raw-signal matching.

Experiments demonstrate zero-shot transfer to real sensor measurements across diverse downstream tasks, including indenter shape identification, force prediction, and shape reconstruction. Furthermore, we show that complementary simulation modalities enrich the latent space in task-specific ways: rigid-body penetration depth aids geometric reasoning, while FEM-based stress fields improve force estimation. By decoupling representations from raw sensor outputs, this aligned latent framework lowers the barrier to efficient sim-to-real transfer, making it uniquely suited for data-intensive paradigms such as large-scale reinforcement learning.

Several directions remain open for future work. First, relaxing the requirement for strictly paired observations across modalities would improve scalability when full simulation coverage is unavailable. Second, extending the approach to dynamic manipulation sequences could leverage the rich information embedded in temporal contact evolution. Finally, as the framework is sensor-agnostic by design, evaluating its transferability across different tactile sensor platforms would further validate its utility as a foundation for scalable tactile-based robot learning.

\endgroup

\small{
\bibliographystyle{IEEEtran}
\bibliography{myrefs}
}

\end{document}